\documentclass[conference]{IEEEtran}
\IEEEoverridecommandlockouts
\usepackage{cite}
\usepackage{amsmath,amssymb,amsfonts}
\pdfoutput=1
\usepackage{algorithmic}
\usepackage{graphicx}
\usepackage{textcomp}
\usepackage{xcolor}
\usepackage{xspace}
\usepackage{xurl}
\def\BibTeX{{\rm B\kern-.05em{\sc i\kern-.025em b}\kern-.08em
 T\kern-.1667em\lower.7ex\hbox{E}\kern-.125emX}}
\usepackage{caption}
\usepackage{mathtools}
\usepackage{multirow}
\usepackage{algorithm}
\usepackage{adjustbox}
\usepackage{tikz}
\usetikzlibrary{positioning}
\usepackage{pgfplots}
\pgfplotsset{compat=1.17}
\usepackage{ctable}

\begin{document}

\title{FlowPrecision: Advancing FPGA-Based Real-Time Fluid Flow Estimation with Linear Quantization}
\author{
 \IEEEauthorblockN{Tianheng Ling, Julian Hoever, Chao Qian, Gregor Schiele}
 \IEEEauthorblockA{Intelligent Embedded Systems Lab, University of Duisburg-Essen, 47057 Duisburg, Germany}
 \IEEEauthorblockA{\{tianheng.ling, chao.qian, gregor.schiele\}@uni-due.de, julian.hoever@stud.uni-due.de}
}


\maketitle
\begin{abstract}

In industrial and environmental monitoring, achieving real-time and precise fluid flow measurement remains a critical challenge. This study applies linear quantization in FPGA-based soft sensors for fluid flow estimation, significantly enhancing Neural Network model precision by overcoming the limitations of traditional fixed-point quantization. Our approach achieves up to a 10.10\% reduction in Mean Squared Error and a notable 9.39\% improvement in inference speed through targeted hardware optimizations. Validated across multiple data sets, our findings demonstrate that the optimized FPGA-based quantized models can provide efficient, accurate real-time inference, offering a viable alternative to cloud-based processing in pervasive autonomous systems.

\end{abstract}


\section{Introduction}

Pervasive in-situ AI systems have immense potential in various applications, such as industrial and environmental monitoring. One critical task in these settings is accurately measuring fluid flow, which is challenging, particularly in harsh environments such as drilling pipes or sewers. Direct-contact physical sensors often degrade quickly and require frequent maintenance, while non-contact sensors may lack precision or may be cost-prohibitive. An emerging solution to this challenge is to use \emph{soft sensors}, which utilize data from cheap, non-contact sensors to estimate flow rate accurately.

Recent advancements in flow estimation have utilized physical formulas, statistical models, and deep learning algorithms. Among them, Tomperi et al.~\cite{tomperi2022estimation} employed multivariate linear regression with cost-efficient distance sensors for wastewater flow estimation. Chhantyal et al.~\cite{chhantyal2016ultrasonic} integrated data fusion with Artificial Neural Networks (ANNs) for ultrasonic level sensor-based flow estimation. Noori et al.~\cite{noori2020non} combined shallow ANNs with radar sensors, and Liang et al.~\cite{liang2023volume} developed models for complex flow scenarios using ultrasonic sensors. However, these approaches primarily focus on cloud-based soft sensors, susceptible to timing and network dependency issues, potentially affecting service reliability.

In response to the need for autonomy and reliability in soft sensor deployment, our study shifts the focus to deploying ANN-based soft sensor algorithms directly on Internet of Things (IoT) devices. This local analysis approach mitigates network dependencies but introduces new challenges due to the limited computational resources available on IoT devices. These constraints necessitate the optimization of ANN models for efficient execution, a process we address through model quantization. Specifically, we propose quantizing ANN models for deployment on embedded Field Programmable Gate Arrays (FPGAs) using linear quantization with adaptive scalers. This method aligns quantization parameters with the actual data distributions of each tensor, thereby enhancing model precision and balancing energy efficiency. Our key contributions include:

\begin{itemize}
    \item We apply linear quantization to Multi-Layer Perceptron (MLP) models for fluid flow estimation, reducing test loss by up to 10.10\%. 
    
    \item We conduct an ablation study to validate our approach against precision loss issues in fixed-point quantization.
    
    \item We test our method across three data sets, demonstrating its broad applicability and robustness.
    
    \item We develop a specialized hardware accelerator that enhances model precision while maintaining energy efficiency and acceptable resource utilization overhead.

\end{itemize}

The reminder content is organized as follows: Section~\ref{sec:related_work} reviews pertinent literature, setting the stage for our research. Section~\ref{sec:fundamentals} delves into the technical aspects of fixed-point and linear quantization. Section~\ref{sec:approach} details our approach and implementation. Section~\ref{sec:experiments} describes the experimental setup, and Section~\ref{sec:results} presents and discusses our findings. Finally, Section~\ref{sec:conclusion} concludes the paper and outlines future research.

\section{Related Work}
\label{sec:related_work}

Accurate flow estimation using soft sensors requires high-bit quantization to ensure precision, mainly due to its regression-oriented nature. Ling et al.~\cite{ling2023autoquitous} advanced in this area by quantizing an MLP model for FPGA platforms, employing a uniform power-of-2 scaling factor in their 8-bit fixed-point quantization approach. While their approach showcased FPGA deployments' speed and energy efficiency, it faltered in precision when compared to linear quantized models on Microcontroller Units (MCUs). The primary limitation of fixed-point quantization stems from its assumption of uniform tensor distribution, leading to notable precision loss amidst real-world data variability. In our pursuit to enhance model precision for soft sensor applications on FPGAs, we also examined the work of Jain et al.~\cite{jain2020trained}. Their introduction of per-tensor scaling considerably ameliorated compatibility across diverse hardware platforms and more effectively adapted to the variability inherent in tensor distributions. Beyond refining the fixed-point quantization, linear quantization is a formidable alternative. This technique applies a zero point and scaling factor to convert floating-point tensors into integer tensors linearly. Gaining traction in prominent deep learning frameworks like TensorFlow and PyTorch, linear quantization is epitomized by the \textit{Brevitas} library~\cite{brevitas} in PyTorch, showcasing enhanced model precision through its application.

Our work builds upon these works, aiming to refine model quantization for FPGA deployments to enhance the precision of soft sensors in flow estimation. We employ adaptive scaling, dynamically adjusting quantization parameters in alignment with the actual data distributions within models. This approach not only overcomes the constraints of traditional power-of-2 scaling but also aims to elevate numerical precision. To counterbalance the computational overhead inherent in adaptive scaling, we incorporated targeted FPGA optimizations, ensuring energy efficiency and performance efficacy.

\section{Fundamentals}
\label{sec:fundamentals}

Quantization in numerical representation involves converting continuous real numbers from the domain \( \mathbb{R} \) into discrete values within the quantized domain \( \mathbb{Q} \). Consider a tensor \( X \), composed of real-valued elements \( x \), and its quantized counterpart \( X_{q} \). As depicted in Equation~\ref{eq:quantization_mapping}, the scale factor \( S \), a floating-point parameter, defines the relationship between \( X \) and \( X_{q} \), while the zero point \( Z \) is a fixed-point parameter representing zero in \( X \). The rounding operation approximates \( x \) to the nearest integer, and the \( \text{clamp} \) function ensures \( x_q \)  remains within the representable range $[q_{\min}, q_{\max}]$ defined by the chosen quantized format, where $q_{\min}$ and $q_{\max}$ denote the lower and upper representable limits of the quantized domain.
De-quantization, detailed in Equation~\ref{eq:dequantization_mapping}, reverses this process, converting \( X_q \) back into an approximate real-valued tensor \( X' \), using the same scale factor \( S \) and zero point \( Z \).


\begin{align}
X \mapsto X_q & = \text{clamp}\left(\text{round}(\frac{X}{S}) + Z, q_{min}, q_{max}\right) \label{eq:quantization_mapping} \\
X_q \mapsto X' & = S \cdot (X_q - Z) \label{eq:dequantization_mapping} 
\end{align}

\subsection{Fixed-point Quantization}

Fixed-point quantization, commonly used in FPGA deployments, is known for simplicity and deterministic nature. It assigns a fixed bit-width to each number, divided into integer and fractional parts. This format is represented as $(a, b)$, where $a$ and $b$ denote bits for fractional values and total bit count, respectively. The scale factor \( S\) is preset to \( 2^{-a} \), with no zero point \( Z \). Despite its utility, fixed-point quantization has limitations. It offers a narrower range than floating-point numbers, risking overflow and precision loss. It may allocate bits inefficiently, covering values outside the needed range and reducing precision. Additionally, it presumes a symmetric data distribution around zero, a condition rarely met in real-world data, leading to quantization errors. These drawbacks highlight the necessity for a more adaptable quantization method to meet varying precision demands in practical applications.
\subsection{Linear Quantization}

In addressing the limitations of fixed-point quantization, our approach utilizes linear quantization with adaptive parameters, as per~\cite{krishnamoorthi2018quantizing}. This method dynamically computes the scaling factors \( S \) and zero points \( Z \) during training, customizing them to each tensor's distribution. Equations~\ref{eq_asymmetric_signed_scale} and~\ref{eq_asymmetric_signed_zero_point} illustrate how these parameters are computed based on the tensor's observed minimum $\alpha$ and maximum $\beta$ values. Here, we adopt a signed integer target with bit-width $b$, where the representable range is defined as $q_{\min} \!=\! -2^{b-1}$ and $q_{\max} \!=\! 2^{b-1} - 1$.  This adaptive approach ensures a more accurate numerical representation, especially for data distributions that are not symmetric around zero, thereby enhancing model performance by minimizing quantization errors and optimizing bit precision utilization.
\begin{align}
S & = \frac{\alpha - \beta}{2^b - 1} \label{eq_asymmetric_signed_scale} \\
Z & = \text{clamp}\left(\text{round}((2^{b-1} - 1) - \frac{\alpha}{S}), -2^{b-1}, 2^{b-1}-1\right) \label{eq_asymmetric_signed_zero_point}
\end{align}

\section{Our Approach}
\label{sec:approach}

Our approach utilizes an MLP model, similar to those in~\cite{noori2020non, ling2023autoquitous}, comprising a hidden layer with a ReLU activation function and an output layer. Adapted for FPGA deployment, our method facilitates more flexible linear quantization within the PyTorch framework. Consistent with the methodologies outlined in~\cite{ling2023autoquitous}, we vary the number of neurons in the hidden layer (10, 30, 60, and 120) to explore the effects of model complexity on performance. Next, we will describe our approach from both software and hardware implementation perspectives.

\subsection{Customized Software Implementation}
\subsubsection{QAT with Linear Quantization}
\begin{figure}[!htb]
    \begin{minipage}{0.25\columnwidth} 
        \centering
        \includegraphics[width=.8\linewidth]{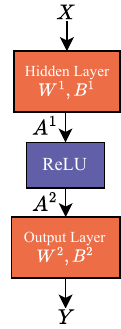}
        \label{fig:model_architecture}
    \end{minipage}
    \hfill
    \begin{minipage}{0.6\columnwidth} 
        \centering
        \footnotesize
        \captionof{table}{Quantization Description}
        \label{tab:quantization_params}
        \begin{tabular}{ccc}
            \specialrule{.15em}{.1em}{.1em}
            \textbf{Layers} & \textbf{\begin{tabular}[c]{@{}c@{}} Quantization \\  Objects\end{tabular}} & \textbf{\begin{tabular}[c]{@{}c@{}}Quantization \\ Parameters\end{tabular}} \\
            \specialrule{.1em}{.05em}{.05em}
             \multirow{4}{*}{\textbf{\begin{tabular}[c]{@{}c@{}}Hidden \\ Layer\end{tabular}}} & $X$ & $S_X$, $Z_X$ \\
                & $W^{1}$ & $S_{W^{1}}$, $Z_{W^{1}}$ \\
                & $B^{1}$ & $S_{B^{1}}$ \\
                & $A^{1}$ & $S_{A^{1}}$, $Z_{A^{1}}$ \\ \hline
             \multirow{2}{*}{\textbf{ReLU}} & $A^{1}$ & $S_{A^{1}}$, $Z_{A^{1}}$ \\
                & $A^{2}$ & $S_{A^{1}}$, $Z_{A^{1}}$ \\ \hline
             \multirow{4}{*}{\textbf{\begin{tabular}[c]{@{}c@{}}Output \\ Layer\end{tabular}}} & $A^{2}$ & $S_{A^{1}}$, $Z_{A^{1}}$ \\
                & $W^{2}$ & $S_{W^{2}}$, $Z_{W^{2}}$ \\
                & $B^{2}$ & $S_{B^{2}}$ \\
                & $Y$ & $S_Y$, $Z_Y$ \\ 
            \specialrule{.15em}{.1em}{.1em}
            \label{tab:quantization_parameters}
            \end{tabular}    
    \end{minipage}
\end{figure}

Quantization-Aware Training (QAT) is a favored quantization strategy that enhances ANNs' resilience to quantization errors. This technique simulates the quantization process during training, incorporating both quantization and de-quantization steps. The non-differentiable nature of rounding operations in backpropagation is addressed using the Straight-Through Estimator~\cite{yin2019understanding}. We transition from the theoretical underpinnings of QAT to its practical application in our model, identifying the primary quantization objects, as detailed in Table~\ref{tab:quantization_parameters}. These include the inputs $X$, weights $W^{1}$, biases $B^{1}$, and activations $A_1$ for the hidden layer, and inputs $A_2$, weights $W_2$, biases $B_2$, and final outputs $Y$ for the output layer. Unlike fixed-point quantization with static scale factors, our approach dynamically calculates the quantization parameters after each training batch, aligning them with the actual data distribution to minimize quantization errors. The specific quantization parameters for each object are concisely summarized in Table~\ref{tab:quantization_parameters}. For example, the quantization scale factor for the hidden layer's inputs is $S_X$, with a corresponding zero point of $Z_X$. Notably, all biases have no zero points for computational simplicity. Although our implementation supports mixed-precision quantization, all quantization objects in this study are uniformly quantized at 8 bits for model consistency and simplified analysis.


In our approach, we adapt the ReLU function's implementation. During QAT, we employ PyTorch's standard \texttt{torch.nn.functional.relu} function. For subsequent integer-only inference, the quantization parameters of the ReLU function's inputs are inherited from the outputs of the preceding layer, i.e., scale $S_{A^{1}}$ and zero point $Z_{A^{1}}$. We use the same quantization parameters for the outputs of the ReLU function to ensure uniformity and coherence in quantization across inputs and outputs. Similarly, for the inputs of the output layer, the inputs $A_2$ inherit their quantization parameters directly from the hidden layer's outputs $A_1$. This approach maintains a consistent and coherent quantization method throughout the model, facilitating the integrity of the quantization process across varying layers.

\subsubsection{Enhanced Integer-only Model Inference}
For deployment on resource-constrained FPGAs, which do not support floating-point operations, we have implemented an integer-only model inference approach. This ensures that all computations and data transfers between layers are conducted using integers, preserving the model's efficiency and precision.

\paragraph{Integer-only Linear Layers}

We employ the integer-only linear layer calculation method proposed by Benoit et al.~\cite{jacob2018quantization}. This method is applied starting from the hidden layer of our model. The computation leverages quantization parameters determined during the QAT phase, which are detailed in Equation~\ref{eq:linear1_step1}. To streamline the calculation, we approximate the bias term \( S_{B^{1}}B^{1}_{q} \) as \(S_{X}S_{W^{1}}B^{1*}_{q} \), leading to the formulation presented in Equation~\ref{eq:linear1_step2}. Notably, the term \( \frac{S_{X}S_{W_{1}}}{S_{A^{1}}} \) in this equation is the sole floating-point component. To maintain our commitment to integer-only computations, we employ bit-shift operations for approximating this term, as elaborated in Equation~\ref{eq:linear1_step4}. Specifically, we first convert the floating-point term into a positive integer \( M_0 \), followed by a right bit-shift operation by \( n \) positions to closely approximate the original floating-point value. This integer-only computation approach is also consistently applied to the output layer, as indicated in Equation~\ref{eq_quant_params_across_layers}, ensuring a uniform integer-only processing methodology throughout our model.

%
\begin{equation}
\footnotesize S_{A^{1}}(A^{1}_{q}-Z_{A^{1}})  \approx S_{X}(X_{q}-Z_{X})S_{W^{1}} (W^{1}_{q}-Z_{W^{1}}) + S_{B^{1}}B^{1}_{q} \label{eq:linear1_step1}
\end{equation}
\begin{equation}
\footnotesize S_{A^{1}}(A^{1}_{q}-Z_{A^{1}})  \approx S_{X}(X_{q}-Z_{X})S_{W^{1}} (W^{1}_{q}-Z_{W^{1}}) + S_{X}S_{W^{1}}B^{1*}_{q} \label{eq:linear1_step2}
\end{equation}
\begin{equation}
\footnotesize A^{1}_{q}  \approx \frac{S_{X}S_{W^{1}}}{S_{A^{1}}}((X_{q}-Z_{X})(W^{1}_{q}-Z_{W^{1}}) + B^{1*}_{q}) + Z_{A^{1}} \label{eq:linear1_step3}
\end{equation}
\begin{equation}
\footnotesize \frac{S_{X}S_{W^{1}}}{S_{A^{1}}} \approx 2^{-n} M_0 \label{eq:linear1_step4}
\end{equation}
\begin{equation}
\footnotesize Y_{q} \approx \frac{S_{A^{1}}S_{W^{2}}}{S_{Y}}((A^{2}_{q}-Z_{A^{1}})(W^{2}_{q}-Z_{W^{2}}) + B^{2*}_{q}) + Z_{Y}
\label{eq_quant_params_across_layers}
\end{equation}

\paragraph{Integer-only ReLU} 
In our implementation, we align with TensorFlow Lite's approach for the integer-only ReLU function. Specifically, we use $Z_{A^{1}}$ as a threshold, as formulated in Equation~\ref{eq_relu_adaptive_quant}. This threshold is designed to align with the zero value in the floating-point space, ensuring that the behavior of our integer-only ReLU function closely mirrors that of its floating-point counterpart. This method is critical in maintaining the precision of the model during quantization.

\begin{equation}
\footnotesize A^{2}_{q} \approx \max(Z_{A^{1}}, A^{1}_{q})
\label{eq_relu_adaptive_quant}
\end{equation}

\subsection{Optimized Model Inference on FPGAs}

To effectively integrate linear quantization within an MLP accelerator, ensuring efficiency on par with fixed-point quantization, we have implemented several key optimizations.

\subsubsection{Linear Layer Optimization}

Our approach to optimizing the linear layer draws inspiration from the designs presented in~\cite{qian2023elasticai}. We have adapted their VHDL templates to meet our specific needs for integer-only inference.

\paragraph{Configurable Parameters} 

The VHDL template in our study incorporates a range of configurable parameters: \( M_{0} \), \( n \), \( Z_X \), \( Z_{W} \), and \( Z_{Y} \). These parameters are crucial for achieving integer-only inference and align closely with those in the software quantization framework, ensuring seamless integration between hardware and software components. Focusing on the output layer as a case in point, the parameters $M_{0}$ and $n$ are instrumental for bit-shifting operations that hone in on the precise approximation of the scale factor $\frac{S_{A^{1}}S_{W^{2}}}{S_{Y}}$. Here, $Z_X$, $Z_{W}$, and $Z_{Y}$ meticulously align with $Z_{A^{1}}$, $Z_{W^{2}}$, and $Z_{Y}$ respectively, thus maintaining the integrity of our quantization approach throughout the computational process.

\paragraph{Pipelined Matrix Multiplication} 
To exploit the parallelization capabilities of FPGAs, we refine our approach for linear quantization and integer-only inference. Our optimizations include a) subtracting zero points before multiplication and accumulation (MAC) operations and b) redesigning the MAC unit with a pipelined architecture to accommodate complex scaling needs. These algorithmic modifications, detailed in Algorithm~\ref{algorithm:enhanced-mac}, ensure efficient parallel task execution, reducing processing delays and enabling higher clock frequencies. Specifically, steps 7 to 9, involving data fetching and zero-point subtraction, are intentionally separated from the main MAC operations. Furthermore, the bit-shifting for approximating \( M \) is strategically placed after the main computation loop, ensuring efficient scaling application just before populating the output buffer \( Y \). Significantly, steps 3 to 5 and 12 to 14 are designed to be executed efficiently within a single 100MHz clock cycle, even though they are in distinct stages.

\begin{algorithm}
    \centering
    \footnotesize
    \caption{MAC Algorithm in the Linear Layer}
    \label{algorithm:enhanced-mac}
    \renewcommand{\algorithmicrequire}{\textbf{Input:}}
    \renewcommand{\algorithmicensure}{\textbf{Output:}}
    \begin{algorithmic}[1]
    \REQUIRE $x$ is an $K$-element vector, $W$ is an $J \times K$ matrix \\
         B is an $J$-element vector
    \STATE Initialization: $sum \leftarrow 0 $, $ j \leftarrow 0$
    \REPEAT
        \STATE $k \leftarrow 0$
        \STATE Load: $W[j][0]$, $x[0]$, $B[j]$
        \STATE $sum \leftarrow sum + B[j]$
            \REPEAT
            \STATE Load $W[j][k+1]$, $x[k+1]$
            \STATE $w \leftarrow W[j][k] - Z_W$, $x \leftarrow x[k] - Z_X$
            \STATE $sum \leftarrow sum + w \cdot x$
            \STATE $k \leftarrow k+1$
            \UNTIL $k=K$ 
        \STATE $y \leftarrow (sum \cdot M_{0}) >> n$
        \STATE Store: $Y[j] \leftarrow y + Z_Y$
        \STATE $j \leftarrow j+1$
    \UNTIL $j=J$ 
    \ENSURE $Y$
    \end{algorithmic}
\end{algorithm}


\subsubsection{ReLU Optimization}
The ReLU function optimization is meticulously calibrated to handle inputs on an element-wise basis, ensuring a straightforward and delay-free operation. Including the configurable parameter \( Z_A \) sets a threshold for the input tensor, enabling the comparator to adjust the output as needed. If the input is less than \( Z_A \), the output is set to this threshold value. Otherwise, it retains the input value. This efficient logic ensures minimal operational delay, allowing immediate output updates following input changes.

\subsubsection{Network Component Integration} 

By integrating the linear layer and ReLU optimizations, we construct a comprehensive network component to realize the MLP model. This component sequentially interconnects layers with activation functions, reflecting the data flow established in the software implementation. The templates for assembling the MLP models are thoroughly validated through ongoing simulations. Our hardware implementation is verified to produce outputs consistent with those from the software implementation, as demonstrated by regular tests conducted on the test data set.

\section{Experiments}
\label{sec:experiments}
\subsection{Data Sets and Task}

\begin{figure}[!htb]
    \centering
    \includegraphics[width=0.8\linewidth]{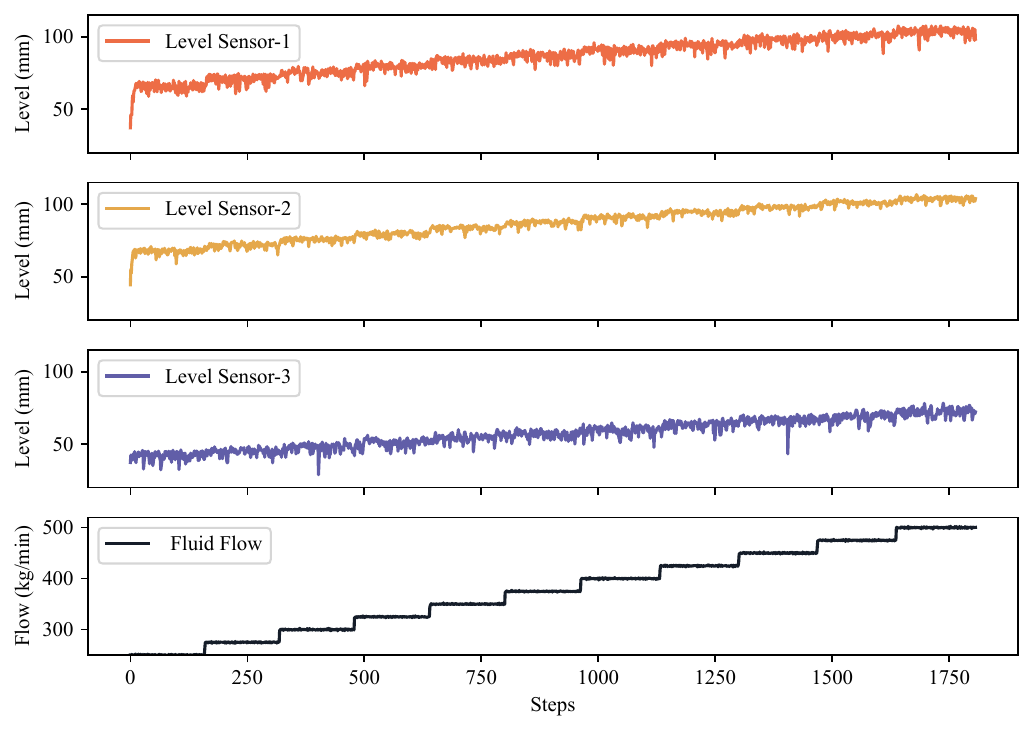}
    \caption{Data Set \textit{DS1}: Measurements With Upward Trend}
    \label{fig:dataset01}
\end{figure}

In this study, we utilized three distinct data sets. The specifics of each data set are summarized in Table~\ref{tab:datasets}. \textit{DS1}, an open data set referenced from~\cite{chhantyal2016ultrasonic}, simulates drilling conditions using an advanced flow loop system. This system includes a mud tank and a Venturi construction, with high-fidelity data captured at a 10kHz sampling rate from three level sensors. The data from these sensors serves as input data, while a precise Coriolis mass flow meter provides the target data, as depicted in Figures~\ref{fig:dataset01}. To further strengthen our model evaluation, we incorporated two additional data sets, \textit{DS2} and \textit{DS3}, generously provided by Viumdal, a co-author of the paper~\cite{chhantyal2016ultrasonic}. These data sets are characterized by upward and downward flow trends, introducing additional complexities and enriching the comprehensiveness of our model's assessment. All three data sets, \textit{DS1}, \textit{DS2}, and \textit{DS3}, are divided using the same proportion: 75\% of the data is allocated for training, and the remaining 25\% is equally split between validation and testing. All data sets were normalized to a range between 0 and 1. To enhance the robustness of our model, we have employed seven-fold cross-validation across all data sets. 


\begin{table}[!htb]
    \centering
    \caption{Flow Estimation Data Sets}
    \setlength{\tabcolsep}{4pt}
    \begin{tabular}{cc}
    \specialrule{.15em}{.1em}{.1em}
    \textbf{Data Sets} & \textbf{Description} \\
    \hline
    \textit{DS1} & 1800 samples with upward trend only \\
    \textit{DS2} & 4439 samples with upward \& downward trends \\
    \textit{DS3} & 4985 samples with upward \& downward trends \\
    \specialrule{.15em}{.1em}{.1em}
    \end{tabular}
    \label{tab:datasets}
\end{table}


\begin{figure*}[!htb]
    \centering
    \includegraphics[width=2\columnwidth]{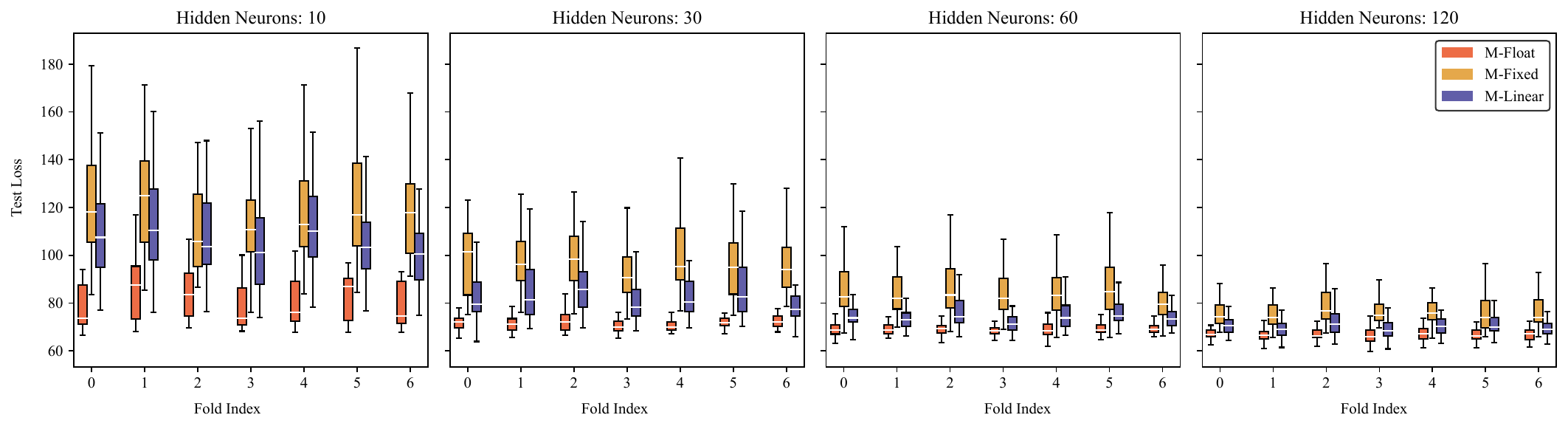}
    \caption{Distribution of Test MSEs Across Various Hidden Neurons}
    \label{fig:exp1}
\end{figure*}

\subsection{Experimental Settings}

Our model training utilized the \textit{Adam} optimizer with hyperparameters set as: \( \beta_1 = 0.9 \), \( \beta_2 = 0.98 \), and \( \epsilon = 10^{-9} \). We started with an initial learning rate of 0.001, employing a scheduler with a step size of 3 and a decay factor \( \gamma \) of 0.5 to adjust it. The Mean Squared Error (MSE) was chosen as the loss function for model training and evaluation. We conducted 100 experiments, each running for 500 epochs and with an early stopping mechanism to avoid overfitting. Notably, the test MSE reported in this study is calculated using de-normalized outputs and target data.


\begin{table}[!htb]
\centering
\renewcommand{\arraystretch}{1.1} 
\caption{Models with Various Configurations}
\setlength{\tabcolsep}{2pt}
\begin{tabular}{ll}
\specialrule{.15em}{.1em}{.1em}
\textbf{Models} & \textbf{Description} \\
\hline
$\textit{M-Float}$ & A 32-Bit Floating-Point Model \\
$\textit{M-Fixed}$ & An (6,8) Fixed-Point Quantized Model Using QAT\\
$\textit{M-Linear}$ & An 8-Bit Integer Quantized Model Using QAT \\
\specialrule{.15em}{.1em}{.1em}
\end{tabular}
\label{tab:models}
\end{table}


To elucidate the distinctions among the models with varying configurations, we present Table~\ref{tab:models} serving as a concise summary. The $\textit{M-Float}$ model, adhering to the standard 32-bit floating-point configuration, closely mirrors the precision levels reported in~\cite{noori2020non}. This model serves as a reference point, helping us gauge the extent of precision loss attributable to quantization. The $\textit{M-Fixed}$ model utilizes a conventional fixed-point (6,8) configuration during QAT, indicating an 8-bit total width with 6 bits allocated for the fractional part. This model is referenced as the baseline from related work~\cite{ling2023autoquitous} for contextual purposes. Furthermore, the $\textit{M-Linear}$ model, employing linear quantization during QAT, is converted into an 8-bit integer format, illustrating our approach to quantization.

\subsection{Target Hardware and Model Deployment}

We deployed the MLP model on a Spartan-7 FPGA for comparison, following the process described in~\cite{ling2023autoquitous}. The process began with converting the trained models within PyTorch framework into VHDL templates. These templates were then tailored to embed the model's weights and biases into ROM instances and integrate other quantization parameters, forming the basis for the RTL designs. Subsequently, these designs were synthesized into bitfiles using the Vivado software suite, enabling deployment on the FPGA. We assessed the model's timing and power consumption using Vivado's analytical tools, which allow for a comparison with the estimates provided in~\cite{ling2023autoquitous}, even without access to their customized hardware.


\section{Results and Evaluation}
\label{sec:results}
\subsection{Test Loss Assessment Across Quantization Methods}

In our first experiment, we conducted a comparative analysis of linear and fixed-point quantization methods, assessing their impact on models with varying complexities. Additionally, the $\textit{M-Float}$ models served as references. This analysis encompassed three model types, each tested with varying numbers of hidden neurons (10, 30, 60, and 120) in the hidden layer, resulting in 12 distinct model variations. Each model underwent 7-fold cross-validation on the \textit{DS1} data set, with each fold involving 100 independent training runs. The distribution of test MSEs is illustrated in Figure~\ref{fig:exp1}. Our findings revealed that the $\textit{M-Float}$ models, featuring floating-point parameters and activations, consistently achieved the lowest test MSEs, surpassing the quantized models. Models with higher counts of hidden neurons tended to yield lower test MSEs. Among the quantized models, the $\textit{M-Linear}$ models generally exhibited lower average and minimum test MSEs compared to the $\textit{M-Fixed}$ models. Moreover, the $\textit{M-Linear}$ models demonstrated greater stability, as indicated by their smaller variance in test MSEs, unlike the $\textit{M-Fixed}$ models. These results, consistent across various folds and model complexities, underscore the effectiveness and reliability of linear quantization for improving model performance.

\subsection{Ablation Study of Linear Quantization}

Following the same experimental setup, we conducted an ablation study on data set \textit{DS1} at Fold 0 to evaluate the impact of linear quantization on model performance. In this study, each linear layer was independently quantized using either fixed-point (F) or linear (L) quantization methods, resulting in 16 distinct model configurations. The performance of these models is summarized in Table~\ref{tab:results_exp2}. When analyzing models with 10 hidden neurons, as shown in the first four rows of Table~\ref{tab:results_exp2}, we observed a consistent increase in MSE when switching from linear to fixed-point quantization. This finding, in line with our initial experiment's results, underscores the critical role of linear quantization in reducing MSE. This pattern holds true for models with a higher number of hidden neurons, further affirming the effectiveness of linear quantization in enhancing model precision across various model sizes.


\begin{table}[!htb]
\centering
\caption{Ablation Study of Linear Quantization}
\begin{tabular}{ccc}
\specialrule{.15em}{.1em}{.1em}
\textbf{\begin{tabular}[c]{@{}c@{}}Hidden \\ Neurons\end{tabular}}& \textbf{\begin{tabular}[c]{@{}c@{}}Quantization Configuration \\ (Hidden Layer / Output Layer)\end{tabular}} & \textbf{MSE} \\ \hline

\multirow{4}{*}{\textbf{10}} 
 & L / L  & \textbf{73.75} \\ 
 & L / F  & 82.00  \\
 & F / L  & 83.37  \\
 & F / F  & 85.86  \\ \hline

\multirow{4}{*}{\textbf{30}} 
& L / L  & \textbf{66.10} \\ 
& L / F  & 74.83  \\
& F / L  & 72.25  \\
& F / F  & 78.36  \\ \hline

 \multirow{4}{*}{\textbf{60}} 
& L / L  & \textbf{59.60} \\ 
& L / F  & 66.36  \\
& F / L  & 67.93  \\
& F / F  & 71.82  \\ \hline

\multirow{4}{*}{\textbf{120}} 
& L / L  & \textbf{62.87} \\ 
& L / F  & 65.38  \\
& F / L  & 67.36 \\
& F / F  & 68.14    \\

\specialrule{.15em}{.1em}{.1em}
\end{tabular}
\label{tab:results_exp2}
\end{table}


\subsection{Stability Assessment on Linear Quantization}

In our further experiments using data sets \textit{DS2} and \textit{DS3}, we adhered to the same experimental setup to assess the generalizability of linear quantization. The summarized outcomes in Table~\ref{tab:results_exp3} consistently highlight the superiority of linear quantization over fixed-point quantization, echoing our earlier observations with data set \textit{DS1}. Particularly noteworthy in data set \textit{DS2} was the up to 9.70\% reduction in test MSE achieved by linear quantization in models with 10 hidden neurons, and a notable decrease of 4.42\% even in the more complex model configuration with 120 hidden neurons. The trends observed in data set \textit{DS3} mirrored these improvements, evident across all neuron configurations, with the largest model configuration exhibiting a 9.40\% reduction in MSE. These consistent results across varied data sets firmly establish the robustness and efficacy of linear quantization, significantly enhancing the performance of quantized models.


\begin{table}[!htb]
\centering
\caption{Model Performance Across Data Sets}
\begin{tabular}{cccccc}
\specialrule{.15em}{.1em}{.1em}
\setlength{\tabcolsep}{8pt}
\multirow{2}{*}{\textbf{\begin{tabular}[c]{@{}c@{}}Data Sets\end{tabular}}} & \multirow{2}{*}{\textbf{\begin{tabular}[c]{@{}c@{}}Hidden \\ Neurons\end{tabular}}} & \multicolumn{3}{c}{\textbf{MSE}}  \\ \cline{3-5}
&  & \textbf{$\textit{M-Float}$} & \textbf{$\textit{M-Fixed}$} & \textbf{$\textit{M-Linear}$} \\ \hline
\multirow{4}{*}{\textit{DS2}}  
& 10 & 75.38 & 82.72 & 74.70($\downarrow$9.70\%)\\
& 30 & 70.74 & 77.30 & 75.55($\downarrow$2.26\%)  \\
& 60 & 67.73 & 74.32 & 72.34($\downarrow$2.66\%)  \\
& 120& 65.51 & 73.09 & 69.86 ($\downarrow$4.42\%)  \\
\hline
\multirow{4}{*}{\textit{DS3}}  
& 10 & 59.49 & 64.41 & 60.03 ($\downarrow$6.80\%) \\
& 30 & 55.53 & 61.14 & 57.63 ($\downarrow$5.74\%) \\
& 60 & 51.17 & 61.68 & 58.22($\downarrow$5.61\%) \\
& 120& 51.60 & 59.24 & 53.67($\downarrow$9.40\%) \\
\specialrule{.15em}{.1em}{.1em}
\end{tabular}
\label{tab:results_exp3}
\end{table}


\subsection{Performance on FPGA Deployment}

Our final set of experiments assessed the performance of models deployed on FPGAs, with a focus on inference time, power consumption, and energy efficiency. As detailed in Table~\ref{tab:results_hw_1}, the $\textit{M-Linear}$ model consistently demonstrated a modest reduction in inference time compared to the $\textit{M-Fixed}$ model, across various hidden neuron configurations on \textit{DS1}. Notably, this increased processing speed was accompanied by an elevated power consumption, often exceeding a 10\% increase. This trade-off between inference speed and power consumption is attributed to the pipelined parallelization in the linear layer implementation. The resource utilization, outlined in Table~\ref{tab:results_hw_2}, shows that the $\textit{M-Linear}$ model requires more Look-Up Tables (LUTs) and Block RAM (BRAM), especially as the neuron count increases. The higher use of Digital Signal Processing slices (DSPs) is necessary for the complex scaling logic, particularly evident at higher model complexities, as detailed in Algorithm~\ref{algorithm:enhanced-mac},  specifically at step 12. The uptick in resource use correlates with the model's complexity, indicating that linear quantization enhances inference speed through more intensive use of FPGA resources. Despite the increased power consumption, the energy cost per inference of the $\textit{M-Linear}$ model is only up to 5.71\% higher than the $\textit{M-Fixed}$ model. Notably, this energy consumption is still quite lower, by 56.8$\times$, compared to model inference on MCUs, as reported in~\cite{ling2023autoquitous}.


\begin{table}[!htb]
\centering
\caption{Model Performance on \textit{XC7S15} FPGA}
\setlength{\tabcolsep}{4pt}
\begin{tabular}{ccccc}
\specialrule{.15em}{.1em}{.1em}
\textbf{\begin{tabular}[c]{@{}c@{}}Models\end{tabular}} & \textbf{\begin{tabular}[c]{@{}c@{}}Hidden \\ Neurons \end{tabular}} & \textbf{\begin{tabular}[c]{@{}c@{}}Inference \\ Time ($\mu$s)\end{tabular}} & \textbf{\begin{tabular}[c]{@{}c@{}}Power \\ (mW) \end{tabular}} & \textbf{\begin{tabular}[c]{@{}c@{}}Energy \\ ($\mu$J/inference)\end{tabular}} \\
\hline
\multirow{4}{*}{\textbf{\begin{tabular}[c]{@{}c@{}}$\textit{M-Fixed}$\end{tabular} }} 
& {10} & 1.04 & 28 & 0.03 \\
& {30} & 3.04 & 29 & 0.09 \\
& {60} & 6.04 & 29 & 0.18 \\
& {120} & 12.04 & 29 & 0.35 \\
\hline
\multirow{4}{*}{\textbf{\begin{tabular}[c]{@{}c@{}}$\textit{M-Linear}$\end{tabular}}} 
& {10} & 1.01 ($\downarrow$2.88\%) & 31 ($\uparrow$10.71\%) & 0.03 (0.0\%) \\
& {30} & 2.81 ($\downarrow$7.57\%) & 32 ($\uparrow$10.34\%) & 0.09 (0.0\%) \\
& {60} & 5.51 ($\downarrow$8.78\%) & 33 ($\uparrow$13.79\%) & 0.18 (0.0\%) \\
& {120} & 10.91 ($\downarrow$9.39\%) & 34 ($\uparrow$17.24\%) & 0.37 ($\uparrow$5.71\%)\\
\specialrule{.15em}{.1em}{.1em}
\end{tabular}
\label{tab:results_hw_1}
\end{table}


\begin{table}[!htb]
\centering
\caption{Resource Utilization on \textit{XC7S15} FPGA}
\setlength{\tabcolsep}{4pt}
\begin{tabular}{ccccc}
\specialrule{.15em}{.1em}{.1em}
\textbf{Models} & \textbf{Hidden Neurons} & \textbf{LUT}(8000) & \textbf{BRAM}(10) & \textbf{DSP}(20) \\
\hline
\multirow{4}{*}{\textbf{\begin{tabular}[c]{@{}c@{}}$\textit{M-Fixed}$\end{tabular}}} 
& {10} & 355 & 0 & 0 \\
& {30} & 373 & 0  & 0  \\
& {60} & 387 & 0.5  & 0  \\
& {120} & 375 & 1  & 0  \\
\hline
\multirow{4}{*}{\textbf{\begin{tabular}[c]{@{}c@{}}$\textit{M-Linear}$\end{tabular}}} 
& {10} & 467  & 0  & 2  \\
& {30} & 463  & 0.5 & 2 \\
& {60} & 479  & 1 & 2 \\
& {120} & 518 & 1.5 & 2 \\
\specialrule{.15em}{.1em}{.1em}
\end{tabular}
\label{tab:results_hw_2}
\end{table}
\vspace{-5mm}


\section{Conclusion and Future Work}
\label{sec:conclusion}

This study has demonstrated the significant advantages of linear quantization in MLP models for FPGA deployment. Our experimental findings have shown that adjusting quantization parameters substantially reduces MSE and improves model stability across various neuron configurations and data sets. While linear quantized models demand increased power consumption and resource utilization, notable gains in inference speed justify these trade-offs, presenting a compelling option for applications where rapid computational responses are essential. Looking forward, we will concentrate on further refining linear quantization techniques. Our primary objective is to limit the precision loss in quantized models to a maximum of 5\% compared to their full-precision counterparts. In addition, we plan to explore quantization at lower bit depths to deploy more complex MLP models for flow estimation on FPGA platforms. This endeavor aims to enhance both the efficiency and effectiveness of these models in real-time flow estimation. 

\noindent{\textbf{Acknowledgments.}} The authors gratefully acknowledge the financial support provided by the Federal Ministry for Economic Affairs and Climate Action of Germany for the RIWWER project (01MD22007C).

\bibliographystyle{IEEEtran}

\bibliography{reference}
\end{document}